# Applying MambaAttention, TabPFN, and TabTransformers to Classify SAE Automation Levels in Crashes

**Shriyank Somvanshi**
Ingram School of Engineering, Texas State University
601 University Drive, San Marcos, Texas 78666
Email: shriyank@txstate.edu

**Anannya Ghosh Tusti**
Ingram School of Engineering, Texas State University
601 University Drive, San Marcos, Texas 78666
Email: gpk30@txstate.edu

**Mahmuda Sultana Mimi**
Ingram School of Engineering, Texas State University
601 University Drive, San Marcos, Texas 78666
E-mail: qnb9@txstate.edu

**Md Monzurul Islam**
Ingram School of Engineering, Texas State University
601 University Drive, San Marcos, Texas 78666
Email: monzurul@txstate.edu

**Sazzad Bin Bashar Polock**
Ingram School of Engineering, Texas State University
601 University Drive, San Marcos, Texas 78666
Email: pay28@txstate.edu

**Anandi Dutta, Ph.D.**
Ingram School of Engineering, Texas State University
601 University Drive, San Marcos, Texas 78666
Email: anandi.dutta@txstate.edu

**Subasish Das, Ph.D.**
Ingram School of Engineering, Texas State University
601 University Drive, San Marcos, Texas 78666
Email: subasish@txstate.edu

**ABSTRACT**

The increasing presence of automated vehicles (AVs) presents new challenges for crash classification and safety analysis. Accurately identifying the SAE automation level involved in each crash is essential to understanding crash dynamics and system accountability. However, existing approaches often overlook automation-specific factors and lack model sophistication to capture distinctions between different SAE levels. To address this gap, this study evaluates the performance of three advanced tabular deep learning models MambaAttention, TabPFN, and TabTransformer for classifying SAE automation levels using structured crash data from Texas (2024), covering 4,649 cases categorized as Assisted Driving (SAE Level 1), Partial Automation (SAE Level 2), and Advanced Automation (SAE Levels 3-5 combined). Following class balancing using SMOTEENN, the models were trained and evaluated on a unified dataset of 7,300 records. MambaAttention demonstrated the highest overall performance (F1-scores: 88% for SAE 1, 97% for SAE 2, and 99% for SAE 3-5), while TabPFN excelled in zero-shot inference with high robustness for rare crash categories. In contrast, TabTransformer underperformed, particularly in detecting Partial Automation crashes (F1-score: 55%), suggesting challenges in modeling shared human-system control dynamics. These results highlight the capability of deep learning models tailored for tabular data to enhance the accuracy and efficiency of automation-level classification. Integrating such models into crash analysis frameworks can support policy development, AV safety evaluation, and regulatory decisions, especially in distinguishing high-risk conditions for mid- and high-level automation technologies.

# INTRODUCTION

Automated vehicles (AVs) are at the forefront of a transformative shift in transportation, promising significant improvements in roadway safety, operational efficiency, and societal mobility. Human error contributes to over 90% of road crashes, positioning AVs as a potentially powerful countermeasure to reduce traffic-related injuries and fatalities (Fagnant and Kockelman, 2015). Indeed, simulations and early real-world studies suggest that higher levels of vehicle automation can lead to significant drops in conflict events and crashes. For instance, high-level (Level 4) self-driving systems have been shown to reduce hazardous driving conflicts by over 90% compared to manual driving in simulation (Wang et al., 2024). Empirical data from test fleets on public roads also show that AVs experience lower crash rates relative to conventional vehicles, with a notable trend of AVs being more frequently rear-ended than acting as the striking vehicle, underscoring a behavioral mismatch between human and automated systems (Kusano et al., 2024).

To standardize the classification of vehicle automation capabilities, the Society of Automotive Engineers (SAE) introduced the J3016 framework, which defines six levels of driving automation (SAE, 2021). Level 0 represents that the human driver performs all aspects of driving. At Level 1, the vehicle can assist with either steering or acceleration/braking (driver assistance) using features like adaptive cruise control. However, the human remains fully responsible, while Level 2 offers combined functions (partial driving automation) but still requires the human driver to monitor the environment. In Level 3 systems (conditional driving automation), vehicles can manage all aspects of driving under some conditions, though a human driver must stand by to intervene upon request. Level 4 vehicles (high driving automation) can operate without human input in certain environments or conditions, and Level 5 denotes a fully AV capable of self-driving in all situations. This taxonomy serves as the global standard for defining and regulating AV capabilities. It is widely used in legal, engineering, and policy frameworks to categorize crash involvement and responsibility based on automation levels (Ding et al., 2024a).

The rise of automation has also accelerated the integration of connected and intelligent systems into modern vehicles. For example, deep learning-based crash severity models using vehicle sensor data have demonstrated over 15% higher accuracy than traditional approaches in real-time risk identification (Rahim and Hassan, 2021). These innovations support improved traffic flow, reduced congestion, and greater mobility for underserved populations through adaptive traffic management. However, these benefits introduce increasing demands for robust methods capable of classifying AV behavior in diverse crash contexts. The increasing presence of automation in vehicles underscores the need for detailed classification frameworks and data-driven modeling tools. These tools are not only essential for post-crash analysis but also pivotal for the development of safety regulations, insurance policies, and public education regarding AV deployment.

**Complexity of Crash Scenarios**

Crashes involving vehicles at different levels of automation introduce a wide range of behavioral and technical variables that complicate causal attribution. For instance, in Level 2 vehicles, the interplay between human oversight and machine operation can lead to driver over-reliance on assistance features, increasing crash risk due to inattention or delayed reaction (Abdel-Aty and Ding, 2024a). In contrast, Level 4 or 5 systems operate independently in defined conditions, making system failure, sensor limitations, or unpredictable edge cases more relevant crash contributors. Accurately identifying the automation level involved in each crash is therefore critical for assessing the underlying cause and for designing appropriate mitigation strategies. Empirical findings highlight how crash risk characteristics diverge across levels: vehicles with Level 3+ automation are involved in significantly fewer crashes than conventional vehicles, except during conditions such as dusk or complex turns (Abdel-Aty and Ding, 2024; Ding et al., 2024; Kusano et al., 2024). Additionally, injury severity outcomes vary with automation levels. Crashes involving highly automated systems (Level 3 and above) have shown lower rates of serious injury, especially when systems operate within their defined parameters (Ding et al., 2024a). However, misclassifying these levels or lacking automation-level metadata can lead to erroneous conclusions in crash causality analysis and policy development. On public roads, AVs in test fleets have exhibited lower overall crash rates than conventional vehicles under similar conditions (Abdel-Aty and Ding, 2024a). Notably, AVs are seldom at fault in the collisions they are involved in and are often struck from behind by human-driven cars, indicating a discrepancy in driving behavior between autonomous and human drivers (Biever et al., 2020). Beyond safety, the societal impact of AVs spans improved mobility for those unable to drive, reduced traffic congestion, and potential environmental gains through smoother traffic flow (Fagnant and Kockelman, 2015).

**Limitations of Current Classification Methods**

Traditional classification methods, which rely on structured reporting forms or rule-based logic, often fail to capture the dynamic complexity of automation-influenced crashes. These methods typically lack the ability to contextualize machine-human interactions, sensor input failures, or partial autonomy transitions. Santos et al. (2022) noted that more than 60% of existing crash severity prediction models omit automation-specific variables, limiting their utility in assessing the evolving risks of automated systems. Moreover, the increasing diversity of vehicle platforms and sensor configurations introduces further inconsistency in how crash data is recorded and analyzed. Recent studies suggest that advanced machine learning models such as tabular deep learning, attention-based transformers, and probabilistic neural networks offer superior performance in uncovering latent patterns in high-dimensional crash data. Basso et al. (2021) demonstrated that vehicle-level risk prediction models using deep learning achieved up to 19% higher classification accuracy compared to traditional logistic regression methods. Similarly, Anik et al. (2024) showed that attention-based transformers achieved an F1 score of 0.87 in predicting conflict events at intersections, underscoring their effectiveness in handling nuanced, time-dependent data.

Given these challenges and opportunities, there is a clear need for more robust, data-driven methods capable of accurately discerning SAE automation levels in crash scenarios. Such techniques will not only enhance crash analysis but also support the broader development of AV safety standards, risk prediction tools, and regulatory frameworks. Unlike most prior research that focused primarily on California DMV's AV crash data, which is characterized by a limited sample size of approximately 750 crashes and lacks information on SAE automation levels, this study offers a novel contribution by utilizing a broader and more representative dataset that includes vehicles classified across the full range of SAE-defined automation levels. By directly linking automation levels, from SAE 1 through 5, with crash severity outcomes, this research provides a more detailed understanding of how different degrees of automation affect crash dynamics. This approach enables a deeper investigation into safety implications associated with transitions in automation, a dimension that has been largely overlooked in previous empirical analyses. In doing so, the study addresses a critical gap by delivering an evidence-based framework for evaluating crash risk and injury severity across the spectrum of vehicle automation, offering valuable insights to support AV policy development, safety design standards, and real-world implementation strategies.

**Research Objectives and Contributions**

Research on classifying vehicle crashes by SAE automation levels is limited, despite the SAE J3016 standard defining levels from no automation (Level 0) to full automation (Level 5). This study addresses this gap by applying advanced deep learning models MambaAttention, TabPFN, and TabTransformer to structured crash data for automation-level classification. These models are chosen for their effectiveness in handling tabular data.

This study aims to: **(1)** assess and compare the classification performance of MambaAttention, TabPFN, and TabTransformer in identifying SAE automation levels using structured crash data attributes.; **(2)** evaluate the models in terms of their suitability, computational efficiency, and interpretability within the context of transportation safety analytics; **(3)** determine the potential of these models to enhance the accuracy, robustness, and operational utility of automation-level classification in crash analysis; and **(4)** provide insights on how automation-level classification can enhance crash investigations, vehicle safety design, and policymaking. Key contributions include a comparative analysis of these deep learning architectures, an evaluation of their applicability in transportation safety, and recommendations for leveraging AI-based classification in AV safety research and deployment.

**Structure of the Paper**

The remainder of this paper is organized as follows. Section 2 reviews the literature on SAE automation levels, their influence on crash severity, and the application of tabular deep learning in transportation safety analytics. Section 3 describes the dataset, preprocessing techniques, and the methodology used to implement and evaluate the deep learning models. Section 4 presents the comparative results of MambaAttention, TabPFN, and TabTransformer, including model accuracy, class-wise performance, and training dynamics. This section also discusses the implications of the findings for crash investigation, AV safety design, and policy development. Finally, Section 5 concludes the paper with a summary of key contributions, study limitations, and directions for future research.

**LITERATURE REVIEW**

**SAE Automation Levels: Definitions and Importance**

Many leading automotive manufacturers, along with several companies outside the traditional automotive sector, are actively engaged in developing vehicles with varying degrees of driving automation (Bachute and Subhedar, 2021). These varying levels are systematically defined in the SAE J3016 standard, published by the Society of Automotive Engineers (SAE) (SAE International, 2021; SAE International and others, 2016). This standard establishes a uniform vocabulary and framework to describe the functionality and scope of automated driving systems, particularly in the context of on-road motor vehicle applications (Tengilimoglu et al., 2023). This comprehensive framework delineates six levels of driving automation, ranging from Level 0 (no automation) to Level 5 (full automation) (SAE International, 2021). Level 1 (Driver Assistance)
5

introduces specific automation features that enable the vehicle to control steering, acceleration, or deceleration based on information about the driving environment. However, the human driver remains responsible for all other aspects of driving and must constantly supervise the system. In level 2 (Partial Automation), the system can manage both steering and acceleration/deceleration simultaneously. Despite this advancement, the human driver must monitor the driving environment and be prepared to intervene immediately if the system fails to respond appropriately.

Level 3 (Conditional Automation) allows the automated driving system to perform all dynamic driving tasks within specific conditions or operational design domains. The human driver is not required to monitor the environment continuously but must be prepared to take over when the system requests intervention. This level marks a significant shift, as the system handles the driving task but relies on the human driver as a fallback. At level 4 (High Automation), the vehicle can perform all driving functions and monitor its environment without human intervention, provided the operational conditions are defined. If the system encounters a situation outside its operational design domain, it can bring the vehicle to a safe stop without requiring human input. This level signifies a substantial advancement in automation, enabling vehicles to operate autonomously in specific scenarios. Level 5 (Full Automation) represents the highest level of driving automation, where the vehicle can perform all driving tasks under all conditions that a human driver could manage. At this level, no human intervention is required, and the vehicle may be designed without traditional driver controls such as a steering wheel or pedals.

Besides categorizing automation levels, the framework outlines several foundational concepts vital for comprehending the architecture and functioning of autonomous systems. These foundational concepts, Dynamic Driving Task (DDT), Automated Driving System (ADS), Operational Design Domain (ODD), and Object and Event Detection and Response (OEDR), are integral to understanding the scope and structure of modern driving automation (Warg et al., 2023). Each plays a crucial role in developing and deploying AV technologies. Together, they support the classification of automation levels, which are visually summarized in the SAE automation levels framework. DDT signifies the real-time, tactical maneuvers needed to navigate a vehicle amidst active traffic (Trypuz et al., 2024). This entails both lateral and longitudinal vehicle control and constant environmental awareness (Parker et al., 2023). DDT does not encompass strategic-level procedures, such as route planning or determining travel destinations. This distinction focuses on the essential functions required for safe, moment-to-moment driving and vehicle operation.

Another key element is the ADS, which comprises a combination of hardware and software capable of performing the entire DDT on a sustained basis. Importantly, this capability can exist within a defined set of constraints known as the ODD, and it applies to vehicles operating at SAE automation levels 3 through 5 (Kaiser et al., 2023; Parker et al., 2023). An ADS must manage all DDT elements autonomously, regardless of the specific driving context in which it operates (SAE International, 2019). The ODD specifies the exact environmental, geographical, temporal, and traffic-related conditions under which an ADS is intended to function safely (Kaiser et al., 2023; SAE International, 2023). For instance, one ADS might be engineered to navigate city streets, while another is optimized for highway driving. The ODD can include road types (e.g., highways vs. local roads), terrain classifications (e.g., mountainous, urban), speed constraints, and environmental conditions such as weather or lighting (Parker et al., 2023). This framework helps delineate the boundaries within which the system is functional and safe (SAE International, 2019). The OEDR function is another critical subcomponent of the DDT (Parker et al., 2023). OEDR refers to the vehicle's ability to perceive its environment by identifying and interpreting various objects and events and then deciding on an appropriate response (Parker et al., 2023; Trypuz et al., 2024; Warg et al., 2023). These could include detecting pedestrians, cyclists, animals, or other vehicles that may threaten safety. The system must perform these tasks reliably within its defined ODD to ensure safe navigation and hazard avoidance (SAE International, 2019). **Figure 1** summarizes the different levels of driving automation according to the SAE standards. The previous standard by NHTSA (2017) categorized automation into simpler classifications, while the SAE framework provides more detailed levels, including distinctions between Driver Assistance, Partial Automation, and Full Self-Driving Automation. The SAE standard is now widely adopted for defining vehicle autonomy, offering a clearer classification for automated driving technologies.

|  | Level 0 | Level 1 | Level 2 | Level 3 | Level 4 | Level 5 |
|---|---|---|---|---|---|---|
| SAE, BASt, OICA | **No Automation** Zero autonomy, the driver performs all driving tasks. | **Driver Assistance** Vehicle is controlled by the driver, but some driving assist features may be included in the vehicle design. | **Partial Automation** Vehicle has combined automated functions, like acceleration and steering, but the driver must remain engaged with the driving task and always monitor the environment. | **Conditional Automation** Driver is a necessity but is not required to monitor the environment. The driver must be ready to always take control of the vehicle with notice. | **High Automation** The vehicle can perform all driving functions under certain conditions. The driver may have the option to control the vehicle. | **Full Automation** The vehicle can perform all driving functions under all conditions. The driver may have the option to control the vehicle. |
| NHTSA* | No Automation | Function Specific Automation | Combined Function Automation | Limited Self-Driving Automation | Full Self-Driving Automation | |

*Previous Standards

**Figure 1. Automation levels (NHTSA, 2017; Zanchin et al., 2017).**

**Factors Affecting the Crash Severity of AV Levels**
Several studies have examined the correlation between SAE automation levels and crash severity. Ding et al. (2024) conducted an exploratory analysis using multi-source data to assess injury severity under different automation levels. Their findings indicated that vehicles equipped with higher automation levels (SAE Level 4) were more likely to be involved in crashes in urban environments, with factors like weather conditions and vehicle mileage influencing injury outcomes. Ren et al. (2022) utilized a hierarchical Bayesian approach to compare crash severity factors between autonomous and conventional driving modes. The study revealed that certain variables, such as turning movements, had divergent effects on crash severity depending on the driving mode. Fu et al. (2025) analyzed real-world AV crash data from the US between 2015 and 2022 to identify key factors affecting crash severity. Their study emphasized the influence of road type, time of day, and vehicle speed on crash outcomes. The study by Abdel-Aty and Ding (2024b) found that while AVs equipped with ADS generally have a lower likelihood of being involved in crashes compared to human-driven vehicles, they exhibit a higher risk of crashes during dawn/dusk conditions and while making turns. Specifically, the odds of an ADS-equipped vehicle being involved in a crash are 5.25 times higher during dawn/dusk and 1.98 times higher during turning maneuvers compared to human-driven vehicles. Ye et al. (2021) observed that injuries to the back, head, and neck were the most prevalent among individuals involved in AV crashes, with AV occupants accounting for approximately 70.83% of those injured.

      Chen et al. (2020) applied an XGBoost model combined with Points of Interest (POI) data to analyze factors affecting the severity of AV crashes. Key contributors to crash severity included weather conditions, vehicle damage degree, accident location, and collision type. Kuo et al. (2024) utilized machine learning models, including random forest and XGBoost, to identify key factors influencing crash severity, such as vehicle manufacturer, damage level, collision type, vehicle movement, involved parties, speed limits, and proximity to points of interest like schools and medical facilities. Chakraborty et al. (2021) applied non-parametric methods, including decision trees and deep neural networks, to classify traffic crash injury severity. Their study highlighted the importance of speed limits and weather conditions in predicting crash outcomes.

Autonomous driving mode has been repeatedly associated with unique risk profiles compared to human-driven modes. Wang and Li (2019) found that crashes occurring while AVs were in autonomous mode had different severity distributions than those driven manually. Khattak et al. (2020) highlighted the influence of disengagement failures, noting that such transitions can elevate crash risk due to driver delay or confusion. Several studies found that the location of AVs during collisions significantly affects severity. Xu et al. (2019) reported that AVs parked on the roadside and those navigating intersections were more likely to be involved in certain crashes. Zhu and Meng (2022) noted increased severity in intersections. Yuan et al. (2022) observed elevated risk in mid-block segments, particularly when other visibility-reducing factors were present. Some studies found that road types affect crash severity in AV-related crashes. Xu et al. (2019) found a connection between one-way roads and certain crash types. Wang and Li (2019) reported that highway crashes involving AVs often involve higher speeds, influencing injury severity. Zhang and Xu (2021) and Yuan et al. (2022) found that nighttime crashes involving AVs were more likely to result in injuries due to reduced visibility and sensor limitations. Das et al. (2020), Ye et al. (2021), and Zhu and Meng (2022) all emphasized that poor lighting or streetlight failure is linked to higher injury severity in AV-involved crashes. Leilabadi and Schmidt (2019) observed that adverse surface conditions (wet, icy roads) raised the likelihood of loss-of-control crashes in AVs. Zhang and Xu (2021) observed that AVs in high-density traffic environments were more prone to severe multi-party crashes due to limited maneuverability and complex scenario interpretation.

Recent studies have highlighted the need for improved data collection and analytical approaches to accurately evaluate crash risks associated with AVs. Scanlon et al. (2024) proposed benchmarks for retrospective automated driving system crash rate analysis using police-reported crash data. They addressed challenges in comparing ADS crash data with human-driven vehicle crash rates, considering factors like underreporting and geographic variability. Zheng et al. (2023) introduced the AVOID dataset, compiling AV operation incident data across the globe. This dataset aims to facilitate research on AV crash analysis and potential risk identification by providing rich samples, diverse data sources, and high data quality. Ye and Lord (2014) compared three commonly used crash severity models, including multinomial logit, ordered probit, and mixed logit, on sample size requirements. Their findings guide the selection of appropriate models for crash severity analysis, particularly in the context of AVs. Lord and Mannering (2010), reviewed statistical methods for analyzing crash-frequency data and assessing various models' suitability for different types of crash data. Their work provides a foundation for selecting appropriate analytical techniques in AV crash studies.

**Tabular Deep Learning for Transportation Safety**

Previous literature has extensively applied tabular deep learning to address transportation safety challenges. One study utilized the TabNet model to analyze pedestrian crash data from Utah (2010–2022), identifying key factors such as pedestrian age, lighting, turning movements, and alcohol involvement, with SHAP enhancing interpretability (Rafe and Singleton, 2024). Two separate studies applied ARM-Net and MambaNet to analyze crash severity among vulnerable road users, child bicyclists and young motorcyclists in Texas and demonstrated strong model performance in predicting fatal and non-injury crashes (Somvanshi et al., 2025b, 2025a). Another study introduced the Feature Group Tabular Transformer (FGTT) model for organizing diverse traffic datasets and improving crash-type prediction (Lares et al., 2024). Additionally, TabPFN, a transformer-based model, has shown high performance in small data settings, requiring no task-specific training (Hollmann et al. 2023). TabPFN performs supervised classification without the need for gradient-based training or hyperparameter tuning, making it suitable for scenarios with limited data. Integration of social traffic data with TabNet has also been shown to enhance traffic event forecasting (Sun et al., 2023) found that integrating social traffic data with the TabNet model enhances traffic event prediction while revealing the importance of external traffic factors. A recent survey comprehensively reviewed deep tabular learning models such as TabNet, SAINT, TabTransformer, MambaNet, and TabPFN, highlighting their architectural innovations and their expanding applicability to structured domains like transportation safety (Somvanshi et al., 2024).

**Review of MambaAttention, TabPFN, and TabTransformer**

To effectively classify the automation level in crash records, robust machine learning methods are needed to handle the structured, tabular crash data. Traditional approaches like gradient-boosted decision trees (e.g., XGBoost) have long dominated tabular data tasks due to their ability to handle mixed data types and missing values. Indeed, past benchmarks showed tree ensembles outperforming early deep learning on typical tabular datasets (Shwartz-Ziv and Armon, 2022). However, recent advances in deep learning are closing this gap by introducing specialized architectures for structured data (Thielmann et al., 2025). Three state-of-the-art models of particular interest are MambaAttention, TabPFN, and TabTransformer. These models employ novel network designs to capture complex patterns in tabular inputs, and they have demonstrated excellent performance on classification tasks that are relevant to transportation safety research. Below, we provide a brief overview of each model's methodology and discuss their applications.

**MambaAttention**

MambaAttention is a novel deep-learning architecture that reinterprets the attention mechanism through a state-space perspective, replacing traditional softmax-based attention with a parameterized convolutional kernel that enables linear-time sequence modeling. Developed as an efficient alternative to Transformers, MambaAttention maintains the ability to capture long-range dependencies by applying learned state-space dynamics to input sequences (Thielmann et al., 2025). Instead of computing pairwise interactions between tokens via dot products, MambaAttention processes each token through a state-space kernel, defined as:

$$\kappa_\theta(t) = Ae^{-Bt} \tag{1}$$

Where, A and B $\in \mathbb{R}^{d \times d}$, are learnable parameters that control temporal decay and memory dynamics. Given an input sequence $X = \{x_1, x_2, \ldots\ldots x_t\}$, where $x_t \in \mathbb{R}^{d_{in}}$, the model first applies learned linear projections to compute token-wise queries, keys, and values: $q_t = W_q x_t$, $k_t = W_k x_t$, and $v_t = W_v x_t$ with $W_q, W_k, W_v \in \mathbb{R}^{d_{in} \times d}$. Rather than aggregating information using global attention weights, the key sequence is convolved with the state-space kernel to produce a memory-aware context vector, $\tilde{k}_t = \sum_{\tau=1}^{t} \kappa_\theta(t - \tau) k_\tau$, which integrates past information with an exponentially decaying influence.

This context vector $\tilde{k}_t$ is then element-wise multiplied with the value stream $v_t$ to produce a gated output, $h_t = \tilde{k}_t \times v_t$, introducing content-sensitive dynamics without relying on attention maps. The gated representation is then projected back to the original input dimension via $y_t = W_0 h_t + x_t$, where $W_0 \in \mathbb{R}^{d \times d_{in}}$, and combined with a residual connection to facilitate gradient flow and maintain training stability. Pre-layer normalization is applied to each block, and multiple MambaAttention layers can be stacked to increase representational depth while preserving scalability.

Experimental results have demonstrated that MambaAttention matches or surpasses Transformer models on various language modeling and sequence tasks while offering greater efficiency and interpretability. Its design facilitates scalability to long sequences and preserves architectural simplicity, making it a strong candidate for structured sequence modeling tasks such as crash report analysis, where understanding temporal or feature interactions is critical and the input features (road conditions, vehicle attributes, driver assistance features present, etc.) form a complex but structured sequence per crash case.

**TabPFN**

TabPFN (Tabular Prior-Data Fitted Network) is a transformer-based foundation model tailored specifically for tabular data, offering exceptional predictive performance with little to no dataset-specific training. Developed by Hollmann et al. (2025), TabPFN adopts a meta-learning approach, enabling it to generalize to new classification tasks without requiring fine-tuning or gradient updates. The model is pre-trained on an extensive collection of synthetic tabular datasets on the order of 100 million tables each simulating a wide array of real-world data-generating processes. Through this large-scale meta-learning strategy, TabPFN effectively learns a universal prior over tabular classification tasks. At the core of TabPFN is its ability to approximate the posterior predictive distribution over class labels:

$$P(y_{test}|D_{train}, x_{test}) \tag{2}$$

Here $D_{train} = \{(x_1, y_1), \ldots\ldots (x_n, y_n)\}$ is the support set and $x_{test}$ is the query input. This predictive distribution is computed through a single forward pass using a Transformer encoder equipped with causal masking, effectively emulating Bayesian inference without any gradient-based optimization. During inference, TabPFN tokenizes each training instance and the test point into a unified sequence and embeds both categorical and numerical features into a shared latent space. Positional encodings are then added to distinguish between training and query points. The sequence is processed by a Transformer to generate a contextualized representation, which is then passed through a softmax layer to produce the final class probabilities. A visual overview of this architecture is shown in **Figure 2**, where each cell in the input table is treated as a node, and 1D feature-wise and sample-wise attention are applied before passing the representation through an MLP to generate predictions (Hollmann et al., 2025).

One of TabPFN's greatest strengths lies in its speed and optimally-tuned performance on small or medium-sized data it often outperforms well-tuned AutoML pipelines yet requires no hyperparameter tuning or iterative gradient updates (Müller et al., 2022). It has also been shown to reach or exceed state-of-the-art performance using only half the usual training data (Hollmann et al., 2025), making it especially well-suited for domains like transportation safety where annotated crash datasets are limited. TabPFN's robustness to outliers, missing values, and structural noise further reinforces its potential as a fast, generalizable, and efficient solution for classifying SAE automation levels from crash reports.

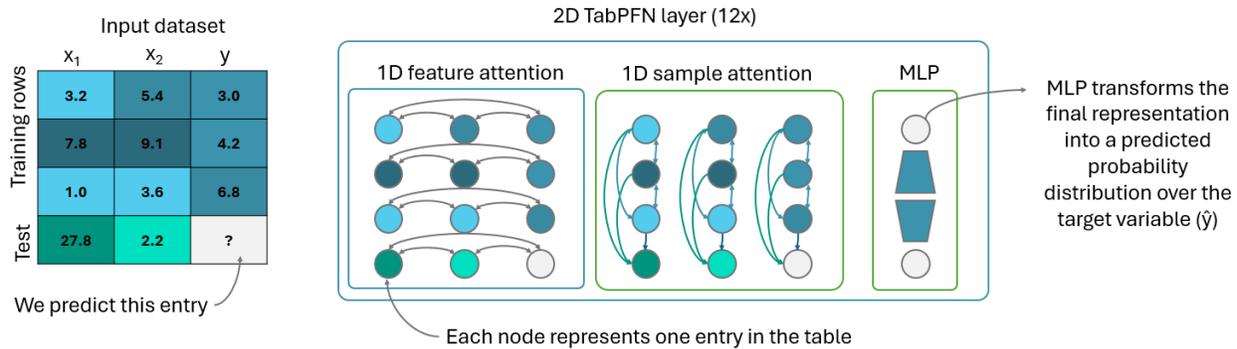

**Figure 2. TabPFN Architecture (Hollmann et al., 2025)**

**TabTransformer**
TabTransformer is a pioneering deep learning model that applies the Transformer architecture originally developed for language processing to structured tabular data (Huang et al., 2020). The key innovation in TabTransformer is its handling of categorical features through contextual embeddings. In crash datasets, many important fields (vehicle make, road type, weather category, etc.) are categorical. TabTransformer encodes each categorical feature value as an embedding vector and uses a multi-head self-attention Transformer encoder to learn contextual embeddings, meaning it learns representations of each feature in the context of other features in the same data record.

The architecture computes these contextual embeddings by first embedding categorical inputs $x_{cat}$ as $E_\phi(x_{cat}) = \{e_{\phi 1}(x_1), \ldots \ldots \ldots e_{\phi m}(x_m)\}$, which are passed through stacked Transformer layers $f_\theta$. The resulting contextualized representations are concatenated with numerical features $x_{count}$, and a multilayer perceptron $g_\psi$ produces the final prediction. The model is trained using the following cross-entropy loss function:

$$\mathcal{L}(x, y) = H(g_\psi \left(f_\theta \left(E_\phi(x_{cat})\right), x_{count}\right), y) \quad (3)$$

This formulation enables the model to capture complex feature dependencies and produce robust predictions even in the presence of missing or noisy inputs. The overall architecture of TabTransformer is illustrated in **Figure 3**. It comprises three key components: (i) a column embedding layer that maps each categorical feature to a dense embedding space, (ii) multiple stacked Transformer layers that model inter-feature interactions via self-attention, and (iii) a multi-layer perceptron that outputs the final prediction. The contextual embeddings generated from the Transformer are concatenated with normalized continuous features and fed into the MLP for supervised learning. This structure enables seamless integration of categorical and numerical data, while leveraging the Transformer's power to capture dependencies among features.

Huang et al. (2020) showed that TabTransformer achieved superior performance to traditional one-hot encoding or target encoding approaches for categorical data, and it outperformed several deep learning baselines, providing on average a 2% AUC uplift over the best previous DNN on tabular benchmarks. Moreover, TabTransformer's accuracy was on par with leading gradient boosting models, and it demonstrated greater robustness to missing or noisy inputs, since the Transformer can capture underlying feature correlations and substitute effectively when some inputs are blank or corrupted. This makes TabTransformer highly applicable to real-world crash data, which often have missing values or reporting inconsistencies.

In terms of applications, TabTransformer and its variants have already appeared in transportation research: for example, (Zhao et al., 2025) integrated an LSTM-TabTransformer model to predict truck travel times in open-pit mining operations, significantly improving prediction accuracy. This success illustrates how combining sequence models (for temporal data) with TabTransformer's tabular attention mechanism can benefit transportation-related tasks. For our problem of classifying SAE automation levels in crashes, TabTransformer offers a way to learn complex relationships between crash circumstances and the presence (or absence) of vehicle automation. By embedding indicators of a vehicle's automation features alongside other crash factors, a TabTransformer-based model could discern subtle patterns (such as particular environmental conditions or driver behaviors that correlate with higher-level automation usage) that simpler models might miss.

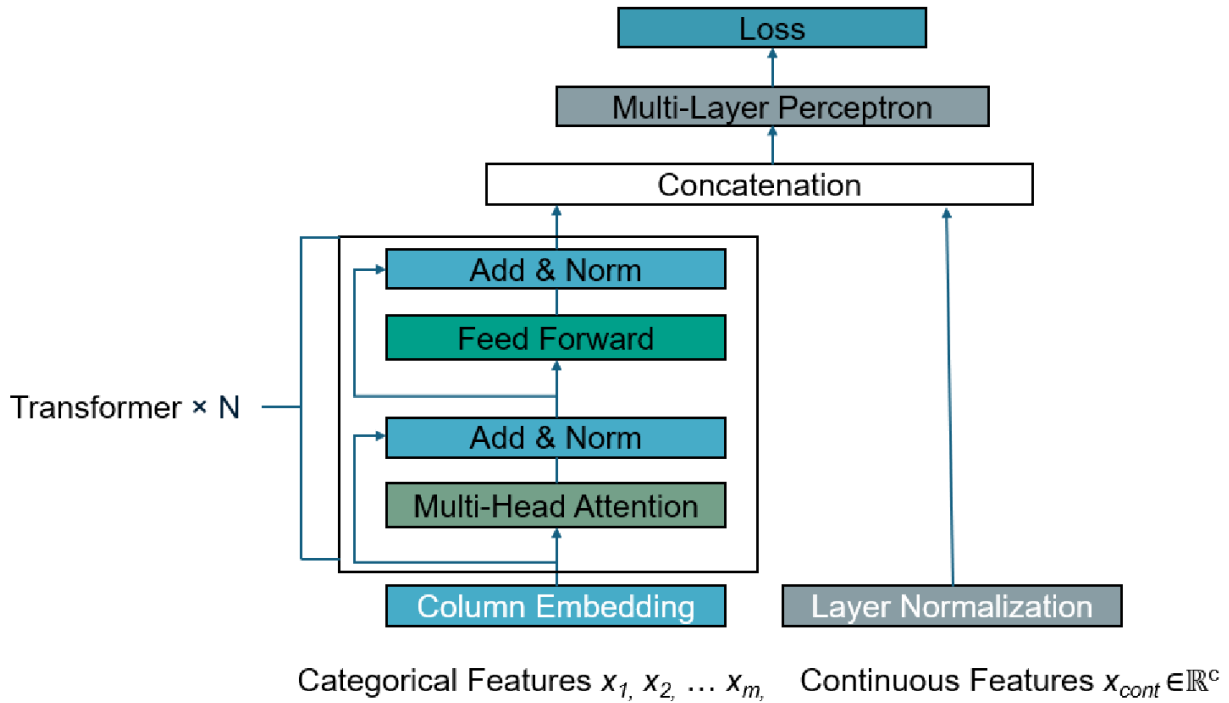

**Figure 3. TabTransformer Architecture (Huang et al., 2020)**

**Summary and Research Gap**
Although the safety performance of AVs has received increasing attention, there is a notable lack of research focused on classifying crashes by SAE automation levels using structured crash datasets. Most existing studies address crash severity or injury outcomes but do not differentiate automation levels, particularly in complex mid-level systems like Partial Automation (SAE Level 2). Traditional statistical and rule-based models often fail to capture the nuanced human-machine interactions and system-level behaviors that define these categories.

Moreover, applications of deep learning in this domain have primarily used generic architectures not tailored for tabular data, overlooking recent advances in models like MambaAttention, TabPFN, and TabTransformer, which are designed specifically for high-dimensional, structured inputs. Prior work has also largely ignored the challenge of accurately identifying Advanced Automation (SAE Levels 3-5) crashes, which are underrepresented and critical for regulatory and policy insights. This study addresses these gaps by applying and evaluating advanced tabular deep learning models to classify automation levels in crashes—an important yet underexplored task in transportation safety analytics.

## DATA AND METHODOLOGY
**Data Description**
This study examines crash data from the state of Texas for the year 2024, focusing specifically on vehicles equipped with various levels of driving automation as defined by the Society of Automotive Engineers (SAE). The dataset, compiled from the Crash Records Information System (CRIS), includes a total of 4,649 crash cases involving vehicles with SAE Level 1 through Level 5 automation. For analysis purposes, the data was categorized into three SAE groups based on sample distribution and automation level characteristics: SAE Level 1 (n = 3,345), SAE Level 2 (n = 1,148), and a combined group of SAE Levels 3 through 5 (n = 156). **To enhance interpretability throughout this paper, we refer to these categories as Assisted Driving (SAE 1), Partial Automation (SAE 2), and Advanced Automation (SAE 3-5 combined), respectively.** The primary target variable in this study is the SAE Automation Level, while the dataset includes a wide range of features characterizing crash, environmental conditions, vehicle types, roadway attributes, and individual demographics as presented in **Table 1**.

A key observation from the dataset is the predominance of crashes occurring under clear weather conditions across all automation categories. Specifically, 75.40% of crashes in the Assisted Driving group, 87.30% in the Partial Automation group, and 82.10% in the Advanced Automation group occurred in clear weather. Similarly, dry road surface conditions were dominant, accounting for 88.00% of crashes in Assisted Driving, 92.00% in Partial Automation, and 91.70% in Advanced Automation. Despite the limited sample size for Advanced Automation, no fatalities were recorded within this group, while fatal injury rates for Assisted Driving and Partial Automation were 0.24% and 0.17%, respectively. Passenger cars were the most common vehicle body style in Advanced Automation crashes (63.50%), which is notably higher compared to 46.00% in Assisted Driving and 43.60% in Partial Automation.

Distinct collision dynamics were observed across automation levels. Fixed-object collisions were considerably more frequent in Advanced Automation crashes (17.90%) compared to Assisted Driving (7.68%) and Partial Automation (6.45%). Population-based differences were also evident: crashes involving Advanced Automation vehicles were more frequently recorded in areas with populations exceeding 250,000 (28.80%) than in Assisted Driving (18.20%) and Partial Automation (11.00%). With respect to posted speed limits, crashes involving Advanced Automation were more likely to occur at higher speeds, with 30.10% occurring at 50-65 MPH. In contrast, crashes in the Assisted Driving category were more evenly distributed across speed limits, with the majority (59.40%) occurring between 30-45 MPH. Regarding injury outcomes, most cases across all automation levels resulted in "not injured" designations, with the highest share observed in Advanced Automation (87.80%). Incapacitating injuries were only reported in the Assisted Driving (0.78%) and Partial Automation (0.52%) groups. Demographic patterns also varied across automation categories. While males were the majority across all groups, their representation decreased from 47.00% in Partial Automation to 40.40% in Advanced Automation. In terms of ethnicity, White drivers accounted for 32.70% of crashes in the Advanced Automation group, compared to 43.90% in Assisted Driving. The most frequently occurring contributing factor across all automation levels was failure to control speed, with the highest incidence observed in Partial Automation (11.10%). While not all variables exhibited statistically significant differences across automation levels, roadway alignment, vehicle body type, and lighting conditions emerged as contextual factors that may interact differently depending on the level of vehicle automation.

**Table 1. Data Description of Key Variables**

| Variable | Assisted Driving | Partial Automation | Advanced Automation | Variable | Assisted Driving | Partial Automation | Advanced Automation |
|---|---|---|---|---|---|---|---|
| | N=3345 | N=1148 | N=156 | | N=3345 | N=1148 | N=156 |
| ***Person Injury Severity (Prsn_Injry_Sev_ID)*** | | | | ***Surface condition (Surf_Cond_ID)*** | | | |
| Incapacitating Injury | 26 (0.78%) | 6 (0.52%) | 0 (0.00%) | Dry | 2943 (88.0%) | 1056 (92.0%) | 143 (91.7%) |
| Killed | 8 (0.24%) | 2 (0.17%) | 0 (0.00%) | Ice | 3 (0.09%) | 0 (0.00%) | 0 (0.00%) |
| Non-Incapacitating Injury | 225 (6.73%) | 51 (4.44%) | 7 (4.49%) | Other | 2 (0.06%) | 2 (0.17%) | 0 (0.00%) |
| Not Injured | 2787 (83.3%) | 989 (86.1%) | 137 (87.8%) | Sand, Mud, Dirt | 4 (0.12%) | 0 (0.00%) | 0 (0.00%) |
| Possible Injury | 299 (8.94%) | 100 (8.71%) | 12 (7.69%) | Standing Water | 21 (0.63%) | 1 (0.09%) | 1 (0.64%) |
| ***Vehicle body (Veh_Body_Styl_ID)*** | | | | Unknown | 4 (0.12%) | 1 (0.09%) | 0 (0.00%) |
| Ambulance | 0 (0.00%) | 1 (0.09%) | 0 (0.00%) | Wet | 368 (11.0%) | 88 (7.67%) | 12 (7.69%) |
| Others | 7 (0.21%) | 3 (0.26%) | 0 (0.00%) | ***Traffic control device (Traffic_Cntl_ID)*** | | | |
| Passenger Car | 1538 (46.0%) | 500 (43.6%) | 99 (63.5%) | Center Stripe/Divider | 156 (5.23%) | 45 (4.54%) | 6 (4.44%) |
| Pickup | 448 (13.4%) | 197 (17.2%) | 16 (10.3%) | Marked Lanes | 1556 (52.1%) | 437 (44.1%) | 79 (58.5%) |
| Police Car/Truck | 26 (0.78%) | 8 (0.70%) | 0 (0.00%) | Others | 196 (6.57%) | 69 (6.96%) | 9 (6.67%) |
| Sport Utility Vehicle | 1222 (36.5%) | 401 (34.9%) | 39 (25.0%) | Signal Light | 644 (21.6%) | 251 (25.3%) | 20 (14.8%) |
| Truck | 25 (0.75%) | 7 (0.61%) | 0 (0.00%) | Stop Sign | 380 (12.7%) | 175 (17.6%) | 21 (15.6%) |
| Truck Tractor | 17 (0.51%) | 3 (0.26%) | 1 (0.64%) | Yield Sign | 52 (1.74%) | 15 (1.51%) | 0 (0.00%) |
| Van | 62 (1.85%) | 28 (2.44%) | 1 (0.64%) | ***Harmful events (Harm_Evnt_ID)*** | | | |
| ***Contributing factor (Contrib_Factr_1_ID)*** | | | | Animal | 82 (2.45%) | 9 (0.78%) | 1 (0.64%) |
| Animal On Road | 85 (2.54%) | 10 (0.87%) | 2 (1.28%) | Fixed Object | 257 (7.68%) | 74 (6.45%) | 28 (17.9%) |
| Changed Lane When Unsafe | 134 (4.01%) | 30 (2.61%) | 4 (2.56%) | Motor Vehicle In Transport | 2806 (83.9%) | 951 (82.8%) | 113 (72.4%) |
| Disregard Stop/Sign/Signal | 121 (3.62%) | 25 (2.18%) | 5 (3.21%) | Other Non Collision | 6 (0.18%) | 0 (0.00%) | 1 (0.64%) |
| Distraction/Inattention/CellPhone Use | 135 (4.04%) | 35 (3.05%) | 8 (5.13%) | Other Object | 17 (0.51%) | 4 (0.35%) | 1 (0.64%) |
| Fail to Control Spd/Unsafe Spd | 287 (8.58%) | 128 (11.1%) | 14 (8.97%) | Overturned | 18 (0.54%) | 5 (0.44%) | 2 (1.28%) |
| Failed To Drive In Single Lane | 73 (2.18%) | 19 (1.66%) | 7 (4.49%) | Parked Car | 125 (3.74%) | 100 (8.71%) | 10 (6.41%) |
| Failed To Yield Right of Way | 295 (8.82%) | 96 (8.36%) | 14 (8.97%) | Pedalcyclist | 15 (0.45%) | 1 (0.09%) | 0 (0.00%) |
| Fatigue/Impair Visi/Under influence of Alc/Drug | 50 (1.49%) | 21 (1.83%) | 5 (3.21%) | Pedestrian | 19 (0.57%) | 4 (0.35%) | 0 (0.00%) |
| Others | 2105 (62.9%) | 762 (66.4%) | 93 (59.6%) | ***First harmful event (FHE_Collsn_ID)*** | | | |
| Turn Improperly/Unsafe | 60 (1.79%) | 22 (1.92%) | 4 (2.56%) | Angle | 807 (24.1%) | 286 (24.9%) | 37 (23.7%) |

| Variable | Assisted Driving | Partial Automation | Advanced Automation | Variable | Assisted Driving | Partial Automation | Advanced Automation |
|---|---|---|---|---|---|---|---|
| *Speed limit (Crash_Speed_Limit)* | | | | One Motor Vehicle | 539 (16.1%) | 197 (17.2%) | 43 (27.6%) |
| 25 MPH or less | 196 (5.86%) | 114 (9.93%) | 11 (7.05%) | Opposite Direction | 325 (9.72%) | 97 (8.45%) | 13 (8.33%) |
| 30-45 MPH | 1988 (59.4%) | 752 (65.5%) | 81 (51.9%) | Others | 38 (1.14%) | 29 (2.53%) | 0 (0.00%) |
| 50-65 MPH | 825 (24.7%) | 201 (17.5%) | 47 (30.1%) | Same Direction | 1636 (48.9%) | 539 (47.0%) | 63 (40.4%) |
| 70 MPH and Over | 336 (10.0%) | 81 (7.06%) | 17 (10.9%) | *Object struck (Obj_Struck_ID)* | | | |
| *Weather condition (Wthr_Cond_ID)* | | | | Barriers and Safety Structures | 120 (3.59%) | 42 (3.66%) | 19 (12.2%) |
| Clear | 2522 (75.4%) | 1002 (87.3%) | 128 (82.1%) | Built Environment Structures | 57 (1.70%) | 13 (1.13%) | 2 (1.28%) |
| Cloudy | 531 (15.9%) | 87 (7.58%) | 16 (10.3%) | Curb/Ditch/Embankment | 47 (1.41%) | 9 (0.78%) | 4 (2.56%) |
| Fog | 19 (0.57%) | 5 (0.44%) | 2 (1.28%) | Hit Tree, Shrub, Landscaping | 35 (1.05%) | 8 (0.70%) | 2 (1.28%) |
| Others | 8 (0.24%) | 1 (0.09%) | 0 (0.00%) | Not Applicable | 2910 (87.0%) | 1033 (90.0%) | 112 (71.8%) |
| Rain | 262 (7.83%) | 52 (4.53%) | 10 (6.41%) | Others | 51 (1.52%) | 7 (0.61%) | 1 (0.64%) |
| Severe Crosswinds | 2 (0.06%) | 0 (0.00%) | 0 (0.00%) | Overturned | 38 (1.14%) | 12 (1.05%) | 6 (3.85%) |
| Sleet/Hail | 0 (0.00%) | 1 (0.09%) | 0 (0.00%) | Traffic Control Devices and Signs | 47 (1.41%) | 16 (1.39%) | 6 (3.85%) |
| Snow | 1 (0.03%) | 0 (0.00%) | 0 (0.00%) | Utility Infrastructure | 40 (1.20%) | 8 (0.70%) | 4 (2.56%) |
| *Lighting condition (Light_Cond_ID)* | | | | *Road class (Road_Cls_ID)* | | | |
| Dark, Lighted | 429 (12.8%) | 209 (18.2%) | 25 (16.0%) | City Street | 1222 (36.5%) | 387 (33.7%) | 50 (32.1%) |
| Dark, Not Lighted | 265 (7.92%) | 64 (5.57%) | 18 (11.5%) | County Road | 165 (4.93%) | 27 (2.35%) | 15 (9.62%) |
| Dark, Unknown Lighting | 17 (0.51%) | 5 (0.44%) | 0 (0.00%) | Farm To Market | 350 (10.5%) | 142 (12.4%) | 13 (8.33%) |
| Dawn | 55 (1.64%) | 12 (1.05%) | 1 (0.64%) | Interstate | 477 (14.3%) | 243 (21.2%) | 34 (21.8%) |
| Daylight | 2515 (75.2%) | 846 (73.7%) | 109 (69.9%) | Non Trafficway | 171 (5.11%) | 130 (11.3%) | 9 (5.77%) |
| Dusk | 60 (1.79%) | 12 (1.05%) | 3 (1.92%) | Other Roads | 0 (0.00%) | 0 (0.00%) | 1 (0.64%) |
| Other (Explain In Narrative) | 1 (0.03%) | 0 (0.00%) | 0 (0.00%) | Tollway | 80 (2.39%) | 11 (0.96%) | 4 (2.56%) |
| Unknown | 3 (0.09%) | 0 (0.00%) | 0 (0.00%) | US & State Hwys | 880 (26.3%) | 208 (18.1%) | 30 (19.2%) |
| *Number of entry roads (Entr_Road_ID)* | | | | *Population group (Pop_Group_ID)* | | | |
| Cloverleaf | 1 (0.03%) | 0 (0.00%) | 0 (0.00%) | 10,000 - 24,999 Pop | 215 (6.43%) | 56 (4.88%) | 18 (11.5%) |
| Five Entering Roads | 4 (0.12%) | 0 (0.00%) | 0 (0.00%) | 100,000 - 249,999 Pop | 813 (24.3%) | 118 (10.3%) | 11 (7.05%) |
| Four Entering Roads | 665 (19.9%) | 232 (20.2%) | 26 (16.7%) | 2,500 - 4,999 Pop | 71 (2.12%) | 10 (0.87%) | 2 (1.28%) |
| Not Applicable | 2188 (65.4%) | 722 (62.9%) | 107 (68.6%) | 25,000 - 49,999 Pop | 301 (9.00%) | 594 (51.7%) | 10 (6.41%) |

| Variable | Assisted Driving | Partial Automation | Advanced Automation | Variable | Assisted Driving | Partial Automation | Advanced Automation |
|---|---|---|---|---|---|---|---|
| Other (Explain In Narrative) | 104 (3.11%) | 24 (2.09%) | 4 (2.56%) | 250,000 Pop And Over | 609 (18.2%) | 126 (11.0%) | 45 (28.8%) |
| Six Entering Roads | 6 (0.18%) | 0 (0.00%) | 0 (0.00%) | 5,000 - 9,999 Pop | 149 (4.45%) | 22 (1.92%) | 5 (3.21%) |
| Three Entering Roads - T | 335 (10.0%) | 157 (13.7%) | 17 (10.9%) | 50,000 - 99,999 Pop | 335 (10.0%) | 51 (4.44%) | 12 (7.69%) |
| Three Entering Roads - Y | 33 (0.99%) | 11 (0.96%) | 2 (1.28%) | Rural | 758 (22.7%) | 158 (13.8%) | 49 (31.4%) |
| Traffic Circle | 9 (0.27%) | 2 (0.17%) | 0 (0.00%) | Town Under 2,499 Pop | 94 (2.81%) | 13 (1.13%) | 4 (2.56%) |
| *Road type (Road_Type_ID)* | | | | *Person age (Prsn_Age)* | | | |
| 2 Lane, 2 Way | 1195 (35.7%) | 366 (31.9%) | 47 (30.1%) | <15 years | 2 (0.06%) | 0 (0.00%) | 0 (0.00%) |
| 4 Or More Lanes, Divided | 721 (21.6%) | 332 (28.9%) | 28 (17.9%) | 15-24 years | 608 (18.2%) | 205 (17.9%) | 28 (17.9%) |
| 4 Or More Lanes, Undivided | 779 (23.3%) | 146 (12.7%) | 49 (31.4%) | 25-34 years | 721 (21.6%) | 259 (22.6%) | 25 (16.0%) |
| Not Applicable | 650 (19.4%) | 304 (26.5%) | 32 (20.5%) | 35-44 years | 693 (20.7%) | 225 (19.6%) | 29 (18.6%) |
| *Road alignment (Road_Algn_ID)* | | | | 45-54 years | 506 (15.1%) | 163 (14.2%) | 19 (12.2%) |
| Curve, Grade | 57 (1.70%) | 6 (0.52%) | 1 (0.64%) | 55-64 years | 339 (10.1%) | 107 (9.32%) | 12 (7.69%) |
| Curve, Hillcrest | 13 (0.39%) | 0 (0.00%) | 1 (0.64%) | 65-74 years | 238 (7.12%) | 60 (5.23%) | 7 (4.49%) |
| Curve, Level | 141 (4.22%) | 22 (1.92%) | 7 (4.49%) | 75+ years | 102 (3.05%) | 41 (3.57%) | 3 (1.92%) |
| Other (Explain In Narrative) | 17 (0.51%) | 11 (0.96%) | 0 (0.00%) | Unknown | 136 (4.07%) | 88 (7.67%) | 33 (21.2%) |
| Straight, Grade | 211 (6.31%) | 41 (3.57%) | 11 (7.05%) | *Ethnicity (Prsn_Ethnicity_ID)* | | | |
| Straight, Hillcrest | 47 (1.41%) | 11 (0.96%) | 2 (1.28%) | Amer. Indian/Alaskan Native | 27 (0.81%) | 4 (0.35%) | 3 (1.92%) |
| Straight, Level | 2858 (85.4%) | 1057 (92.1%) | 134 (85.9%) | Asian | 269 (8.04%) | 45 (3.92%) | 14 (8.97%) |
| Unknown | 1 (0.03%) | 0 (0.00%) | 0 (0.00%) | Black | 446 (13.3%) | 70 (6.10%) | 18 (11.5%) |
| *Intersection related (Intrsct_Relat_ID)* | | | | Hispanic | 986 (29.5%) | 629 (54.8%) | 36 (23.1%) |
| Driveway Access | 371 (11.1%) | 127 (11.1%) | 15 (9.62%) | Other | 20 (0.60%) | 5 (0.44%) | 0 (0.00%) |
| Intersection | 811 (24.2%) | 258 (22.5%) | 34 (21.8%) | Unknown | 128 (3.83%) | 83 (7.23%) | 34 (21.8%) |
| Intersection Related | 621 (18.6%) | 274 (23.9%) | 19 (12.2%) | White | 1469 (43.9%) | 312 (27.2%) | 51 (32.7%) |
| Non Intersection | 1542 (46.1%) | 489 (42.6%) | 88 (56.4%) | *Gender (Prsn_Gndr_ID)* | | | |
| | | | | Female | 1706 (51.0%) | 529 (46.1%) | 61 (39.1%) |
| | | | | Male | 1532 (45.8%) | 540 (47.0%) | 63 (40.4%) |
| | | | | Unknown | 107 (3.20%) | 79 (6.88%) | 32 (20.5%) |

*Note here: Assisted Driving = SAE level 1; Partial Automation = SAE level 2; and Advanced Automation = SAE level 3-5 combined

**Methodology**

The overall structure of the study is illustrated in **Figure 4**, which outlines the two-stage workflow adopted for this research. The first stage involves data preparation, where raw crash records are cleaned, encoded, and balanced using SMOTEENN to address class distribution issues. In the second stage, the resampled data is used for SAE automation level classification. The dataset is split into 60% for training, 20% for validation, and 20% for testing. Three deep learning models, MambaAttention, TabTransformer, and TabPFN, are then applied to predict SAE levels, categorized into three groups: Assisted Driving, Partial Automation, and Advanced Automation. This flowchart serves as a visual summary of the methodology, providing a holistic view of the process that underpins the analyses presented in the subsequent sections.

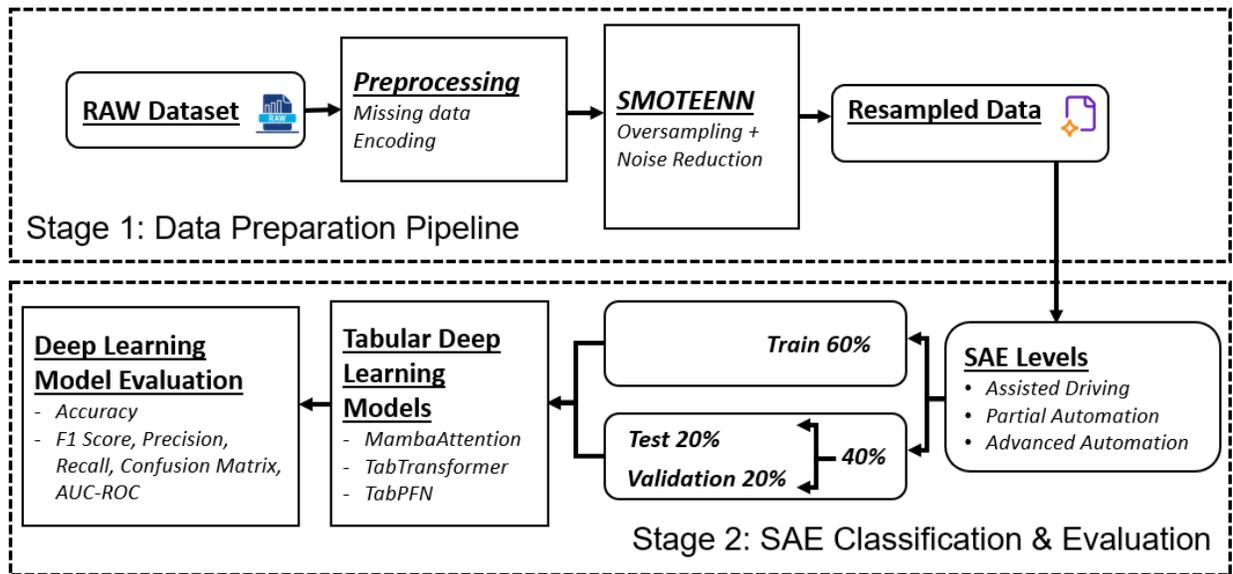

**Figure 4: Flowchart of Combined Study**

One critical challenge encountered in this study was the pronounced class imbalance in the target variable, which classifies crash events into three levels of automation: Assisted Driving, Partial Automation, and Advanced Automation. To better understand the sequential nature of crash events that culminate in these automation levels, we developed a Sankey diagram (**Figure 5**) mapping the flow from crash context through road type, collision type, and injury severity to the engaged AV level. This visualization revealed a highly imbalanced structure, with most crash sequences terminating in Assisted Driving and Partial Automation. In contrast, very few flows reached Advanced Automation, indicating that higher automation levels are associated with rare or underreported crash patterns.

This imbalance was not limited to the final classification but was compounded throughout the sequence of contributing factors. For example, crashes occurring at intersections or involving incapacitating injuries were already underrepresented in the dataset; when these scenarios flowed into the Advanced Automation category, their scarcity became even more pronounced. This finding illustrates that the class imbalance was not merely a label-level issue but a structural one embedded in the data's sequential logic.

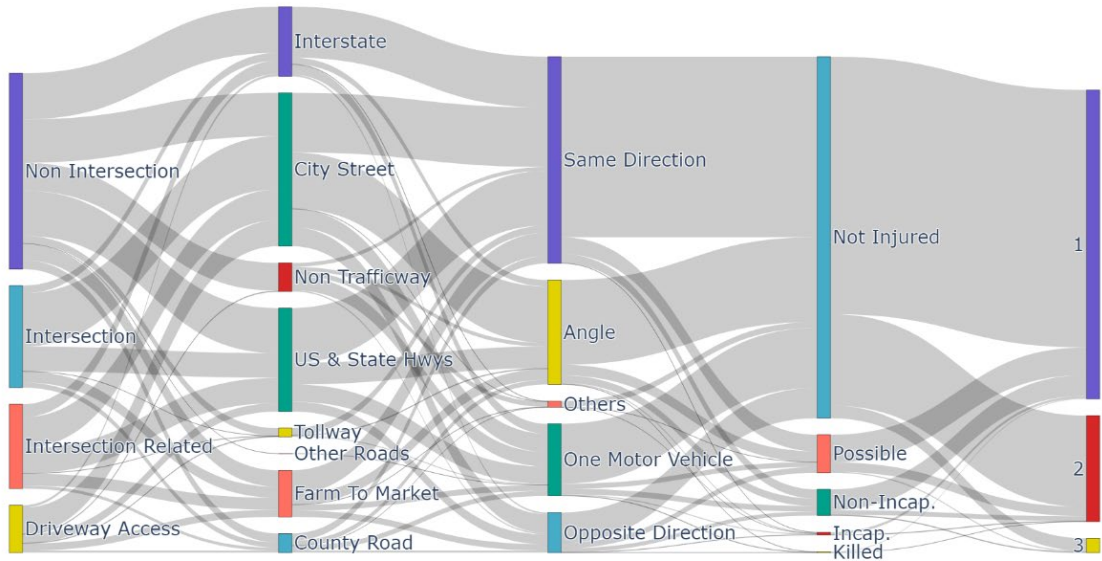

**Figure 5** Sankey diagram showing the flow of crash characteristics from context to SAE automation level. Link width reflects the frequency of transitions, highlighting dominant pathways and class imbalance across stages (Here 1, 2, and 3 denote Assisted Driving, Partial Automation, and Advanced Automation, respectively)

Such a layered imbalance necessitates not only robust classification models (e.g., MambaAttention, TabPFN) but also appropriate data-level balancing techniques. To address this, we applied SMOTEENN, a hybrid resampling strategy (Hairani and Priyanto, 2023) that combines Synthetic Minority Over-sampling (SMOTE) with Edited Nearest Neighbors (ENN) cleaning. This procedure increased the total sample size to 7,300, augmenting the dataset by 983 samples for the Assisted Driving level, 2,972 for the Partial Automation level, and 3,345 for the Advanced Automation level. This method allowed us to both amplify underrepresented Advanced Automation samples and remove noisy or overlapping instances, enhancing the stability and reliability of the classifiers across all automation levels. This resampling strategy not only improved data representation for rare categories but also enhanced the models' capacity for generalization.

The effects of the SMOTEENN resampling strategy on the feature distribution are illustrated in **Figure 6**, which presents kernel density estimates (KDEs) for the variable Traffic control device-signal light (Traffic_Cntl_ID_Signal Light). **Figure 6(a)** compares the overall distributions of this feature in the original versus resampled datasets, revealing that while SMOTEENN significantly increased the number of samples, it preserved the overall shape of the data distribution. **Figure 6(b)** provides class-wise KDE plots across the three SAE levels: Assisted Driving, Partial Automation, and Advanced Automation before and after resampling. These plots demonstrate that the resampling procedure not only improved class representation but also maintained the general structure of the underlying data across automation levels. This balance between augmentation and fidelity is critical for ensuring that synthetic samples introduced via SMOTEENN do not distort key feature distributions, particularly in safety-sensitive domains such as AV crash analysis.

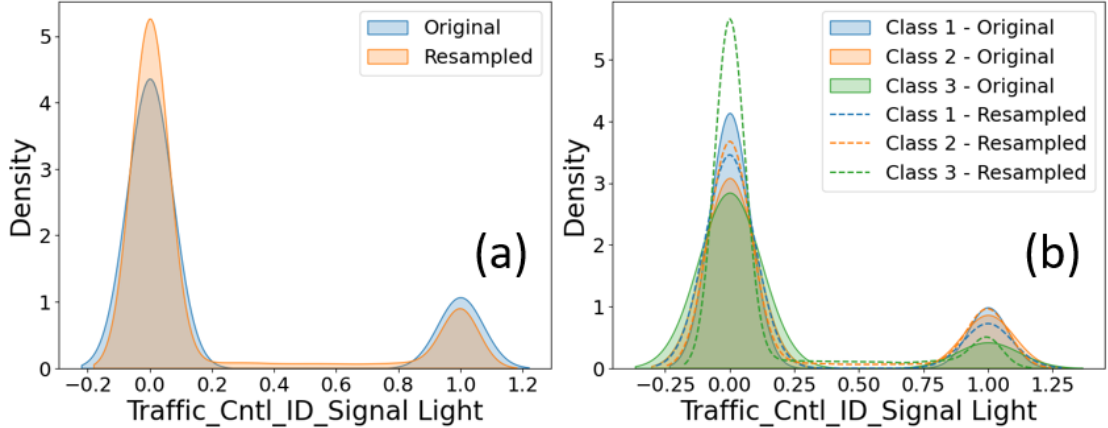

**Figure 6. Kernel density estimates for the feature "Traffic control device-signal light" (Traffic_Cntl_ID_Signal Light) before and after SMOTEENN resampling. (a) Overall distribution comparing original and resampled datasets, (b) Class-wise KDEs for SAE Levels** (Note here: 1 = Assisted Driving; 2 = Partial Automation; 3 = Advanced Automation)

**Hyperparameter Tuning**

To ensure fair and effective model comparison, appropriate hyperparameter configurations were applied for each of the three deep learning models used in this study: MambaAttention, TabTransformer, and TabPFN (**Table 2**). Hyperparameter tuning was conducted using established best practices, with settings informed by prior literature and empirical validation where applicable. For the MambaAttention model, a hidden dimension of 256 was used along with 8 attention heads and a dropout rate of 0.3 to prevent overfitting. The model was optimized using the Adam optimizer with a learning rate of 0.001 and a weight decay of 0.0005 to regularize the network. A learning rate scheduler (StepLR) was applied with a step size of 10 and a gamma value of 0.5 to reduce the learning rate during training. The model was trained for 50 epochs with a batch size of 32.

For the TabTransformer model, the embedding dimension was set to 64 with 4 attention heads and 2 transformer layers. The feedforward layer was configured with a dimensionality of 128, and a dropout rate of 0.1 was applied. Similar to MambaAttention, the model used the Adam optimizer with a learning rate of 0.001, weight decay of 0.0005, and StepLR scheduler with a step size of 10 and gamma of 0.5. Training was conducted for 50 epochs using a batch size of 32. In contrast, the TabPFN model required no manual hyperparameter tuning. TabPFN is designed as a zero-shot inference model, employing a pretrained transformer architecture trained on a wide meta-distribution of tabular tasks. For deployment, the model was configured with the device set to CUDA for GPU acceleration and a fixed random seed of 42 to ensure reproducibility. The consistency in training setup across models, particularly in terms of batch size, number of epochs, and learning rate schedule (where applicable), was maintained to ensure a fair evaluation of model performance in classifying SAE automation levels.

**Table 2. Hyperparameter Tuning**

| Model | Hyper Parameters |
|---|---|
| MambaAttention | Hidden dimension of 256, dropout rate of 0.3, 8 attention heads, Adam optimizer, learning rate of 0.001, weight decay of 0.0005, StepLR scheduler with step size 10 and gamma 0.5, batch size of 32, and trained for 50 epochs. |
| TabTransformer | Embedding dimension of 64, 4 attention heads, 2 transformer layers, feedforward dimension of 128, dropout rate of 0.1, Adam optimizer, learning rate of 0.001, weight decay of 0.0005, StepLR scheduler with step size 10 and gamma 0.5, batch size of 32, and trained for 50 epochs. |
| TabPFN | No hyperparameter tuning was required. The model uses a pretrained zero-shot inference approach, configured with device set to CUDA and random state of 42. |

# RESULTS

## Validation of Experiments

To evaluate the effectiveness of deep learning models in classifying SAE automation levels from structured crash data, three models were implemented: MambaAttention, TabTransformer, and TabPFN. Each model was trained and validated on a resampled dataset comprising 7,300 crash records generated using the SMOTEENN technique. This balanced dataset consisted of 983 samples labeled as Assisted Driving, 2,972 as Partial Automation, and 3,345 as Advanced Automation, as detailed in **Table 3**.

Among the three models, MambaAttention achieved the highest classification accuracy, followed closely by TabPFN, while TabTransformer exhibited comparatively lower performance. MambaAttention was trained for 50 epochs with early stopping and leveraged state-space modeling with efficient attention mechanisms, which likely contributed to its superior performance on this structured, multi-class classification task. TabPFN, in contrast, employed a pretrained zero-shot inference strategy, meaning it was trained beforehand on a wide variety of synthetic tabular tasks and can generalize to new datasets without any additional model tuning or retraining. This allows TabPFN to make reliable predictions directly on unseen tabular data, offering both speed and adaptability while avoiding overfitting.

TabTransformer, despite its architectural advantages for handling categorical data, lagged behind the other two models with an accuracy of 63%. This reduced performance can potentially be attributed to a few key factors. First, TabTransformer's reliance on embedding transformations and multi-head attention over categorical features may be less effective when the number of high-cardinality categorical variables is limited, as is the case in many transportation safety datasets. Second, its architecture may be more sensitive to feature sparsity and subtle interactions between numerical and categorical fields, which are better captured by MambaAttention's dynamic attention structure and TabPFN's meta-learned representations. Furthermore, TabTransformer requires fine-tuned embeddings and sufficient training epochs to stabilize its performance, which could have limited its ability to generalize under the resampled training conditions.

**Table 3 Model Accuracy, Training Details, and Class-wise Sample Distribution Post-Resampling**

| Model | Accuracy (%) | Epochs | Number of Samples | | |
|---|---|---|---|---|---|
| | | | Assisted Driving | Partial Automation | Advanced Automation |
| MambaAttention | 97 | 50 (Early Stopping) | 983 | 2,972 | 3,345 |
| TabTransformer | 63 | | 983 | 2,972 | 3,345 |

| Model | Accuracy (%) | Epochs | Number of Samples | | |
|---|---|---|---|---|---|
| | | | Assisted Driving | Partial Automation | Advanced Automation |
| TabPFN | 93 | Pretrained (Zero-Shot) | 983 | 2,972 | 3,345 |

**Model Performance**

To assess the effectiveness of each model in classifying crash events across the three SAE automation levels, a detailed evaluation was conducted using precision, recall, F1-score, and accuracy as performance metrics. The results are summarized in **Table 4**. The MambaAttention model consistently outperformed the other models across all SAE categories. It achieved an overall accuracy of 97% or higher across all automation levels, with an F1-score of 88% for Assisted Driving, 97% for Partial Automation, and 99% for Advanced Automation. Notably, the model achieved perfect recall and accuracy for the Advanced Automation category, indicating its robustness in identifying cases that often suffer from data sparsity. These results suggest that MambaAttention was highly effective in learning representative patterns from the balanced dataset, particularly excelling in detecting more complex crash scenarios linked to higher automation levels.

TabPFN also delivered a strong performance, achieving 96% accuracy in the Assisted Driving category and 97% in Advanced Automation. It maintained high precision across all classes, with an especially strong performance in identifying Advanced Automation events (99% precision and 98% recall), resulting in an F1-score of 99%. However, its performance was somewhat lower in the Partial Automation category, where its recall dropped to 88%, suggesting that while the model was precise, it occasionally missed cases from this mid-level class.

In contrast, TabTransformer showed comparatively weaker performance, particularly for lower automation levels. While it achieved 84% accuracy for Assisted Driving, it exhibited the lowest performance on Partial Automation, with only 50% accuracy and an F1-score of 55%. Its recall scores were also notably lower, indicating that the model often failed to detect true instances from this class. This performance gap could be attributed to TabTransformer's sensitivity to feature representation and limited capacity to model complex interactions in tabular data when categorical feature richness is low.

The superior performance of MambaAttention and TabPFN in accurately classifying crash events across SAE automation levels holds significant implications for transportation safety and policy. Notably, their high recall and F1-scores in identifying Advanced Automation crashes are crucial, given the rarity and underreporting of such incidents. Accurate classification enhances our understanding of these high-level automation crashes, which is vital for developing targeted safety measures and informing regulatory frameworks. Conversely, the lower performance of TabTransformer, particularly in the Partial Automation category, underscores the challenges in detecting and analyzing crashes involving systems where human drivers share control with automation. This is especially pertinent as SAE Level 2 systems are prevalent on roads today, and misclassification can impede the development of effective safety interventions.

**Table 4. Class-wise Performance Metrics for SAE Automation Level Classification**

| Model | Category | Precision (%) | Recall (%) | F-1 Score (%) | Accuracy (%) |
|---|---|---|---|---|---|
| MambaAttention | Assisted Driving | 94 | 84 | 88 | 83 |
|  | Partial Automation | 96 | 97 | 97 | 97 |
|  | Advanced Automation | 98 | 100 | 99 | 100 |
| TabTransformer | Assisted Driving | 42 | 84 | 56 | 84 |
|  | Partial Automation | 61 | 50 | 55 | 50 |
|  | Advanced Automation | 79 | 68 | 73 | 67 |
| TabPFN | Assisted Driving | 69 | 97 | 81 | 96 |
|  | Partial Automation | 99 | 88 | 93 | 87 |
|  | Advanced Automation | 99 | 98 | 99 | 97 |

To further evaluate the classification behavior of the models, confusion matrices were generated for MambaAttention, TabTransformer, and TabPFN, as shown in Figure 7. These matrices provide a detailed view of class-wise prediction accuracy and the nature of misclassifications across the three SAE automation levels, Assisted Driving, Partial Automation, and Advanced Automation. The confusion matrix for MambaAttention (**Figure 7a**) demonstrates highly consistent performance across all categories. Notably, the model achieved perfect classification for Advanced Automation, with all 669 samples correctly identified. Misclassifications for the Assisted Driving and Partial Automation categories were minimal and largely occurred between adjacent SAE levels, reflecting the model's strong discriminative capability even in potentially ambiguous cases. These results align with its high precision and recall scores reported earlier and underscore its suitability for real-world crash classification tasks involving high-stakes automation scenarios.

In contrast, TabTransformer (**Figure 7b**) exhibited substantial confusion between Partial Automation and the neighboring SAE levels. Of particular concern is the misclassification of 176 Partial Automation samples as Assisted Driving and 121 as Advanced Automation, resulting in both high false positive and false negative rates. This reinforces the model's lower performance on mid-level automation categories observed in the previous evaluation metrics. Such errors are critical, as SAE Level 2 systems currently among the most widely deployed automation technologies require accurate differentiation due to their hybrid operational nature, where both human and system share control (Cooper et al., 2023). Misclassifying these cases can adversely impact post-crash analysis and undermine targeted policy recommendations (National Transportation Safety Board, 2018).

The confusion matrix for TabPFN (**Figure 7c**) indicates well-balanced performance, with accurate classification across all automation levels. The model correctly identified 629 Advanced Automation samples and 193 Assisted Driving cases, with fewer misclassifications than TabTransformer in the Partial Automation category. While a moderate number of Partial Automation cases were misclassified as Assisted Driving (74 instances), TabPFN maintained strong recall and precision, benefiting from its meta-trained architecture and generalization capacity under zero-shot inference.

These class-wise patterns validate the overall superiority of MambaAttention and TabPFN in modeling complex automation-level distinctions within imbalanced crash datasets. Importantly, their ability to reduce misclassification in both lower (Assisted Driving) and higher (Advanced Automation) automation levels strengthens the reliability of crash severity analysis, enhances SAE-specific data reporting accuracy, and supports informed decision-making in the development of safety regulations for increasingly AV fleets.

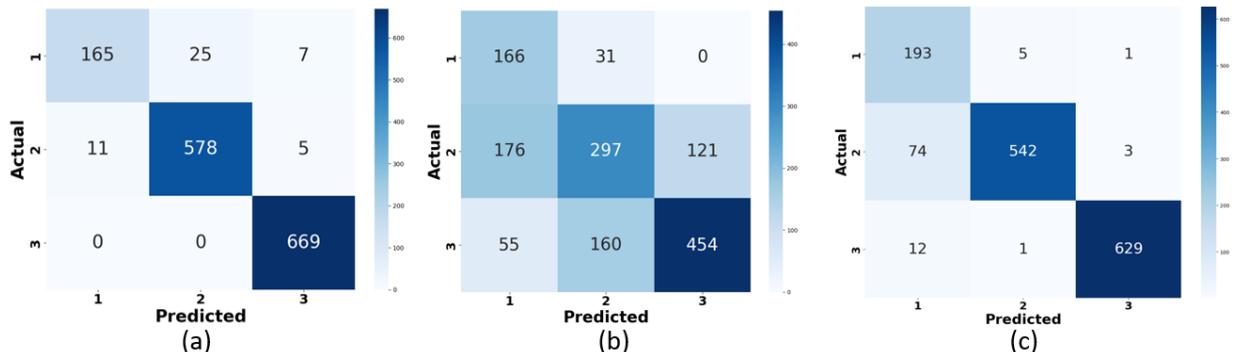

**Figure 7. Confusion Matrix of (a) MambaAttention, (b) TabTranformer, (c) TabPFN**
(Here: 1 = Assisted Driving; 2 = Partial Automation; 3 = Advanced Automation)

**Figure 8** and **Figure 9** illustrate the training dynamics of the MambaAttention and TabTransformer models, respectively, over 50 epochs, including both loss and accuracy curves for training and validation subsets. These plots offer important insights into each model's learning behavior, convergence stability, and generalization capacity. In **Figure 8**, the MambaAttention model demonstrates stable and efficient learning. The training and validation loss curves (**Figure 8a**) show a consistent downward trend, converging smoothly after approximately 30 epochs. Similarly, the accuracy curves (**Figure 8b**) reveal rapid performance gains during the initial epochs, followed by steady improvement and stabilization above 95% validation accuracy, with minimal variance between training and validation. This indicates not only strong convergence but also effective generalization to unseen data, further reflected in its superior overall performance metrics.

In contrast, **Figure 9** shows the training behavior of the TabTransformer model. While the loss curve (**Figure 9a**) does exhibit a general downward trend, the validation loss remains notably more erratic compared to MambaAttention. This suggests greater sensitivity to noise or potentially suboptimal feature interaction modeling. The validation accuracy curve (**Figure 9b**) also fluctuates more prominently and saturates around 65%, indicating underfitting or limited representational capacity in capturing the nuanced distinctions across SAE levels, particularly for mid-level automation (Partial Automation). These patterns are consistent with TabTransformer's lower recall, and F1-score observed in earlier sections.

It is important to note that no training loss or accuracy plots are provided for the TabPFN model. This is because TabPFN operates in a zero-shot inference mode and does not involve task-specific training. It leverages a pretrained transformer model that has been meta-learned on a broad distribution of synthetic tabular tasks and makes predictions without any fine-tuning or gradient-based optimization during deployment. As such, traditional epoch-wise training curves are not applicable, and the model instead provides direct predictions upon ingestion of the dataset.

Taken together, the training performance visualizations further reinforce the earlier findings. MambaAttention exhibits fast and stable convergence with minimal overfitting, aligning with its high predictive accuracy and class-wise robustness. Meanwhile, TabTransformer's fluctuating learning trajectory reflects the challenges it faces in generalization, supporting its relatively weaker performance in this study. TabPFN's design bypasses traditional training altogether, positioning it as an efficient alternative for rapid inference when pretraining aligns well with target task characteristics.

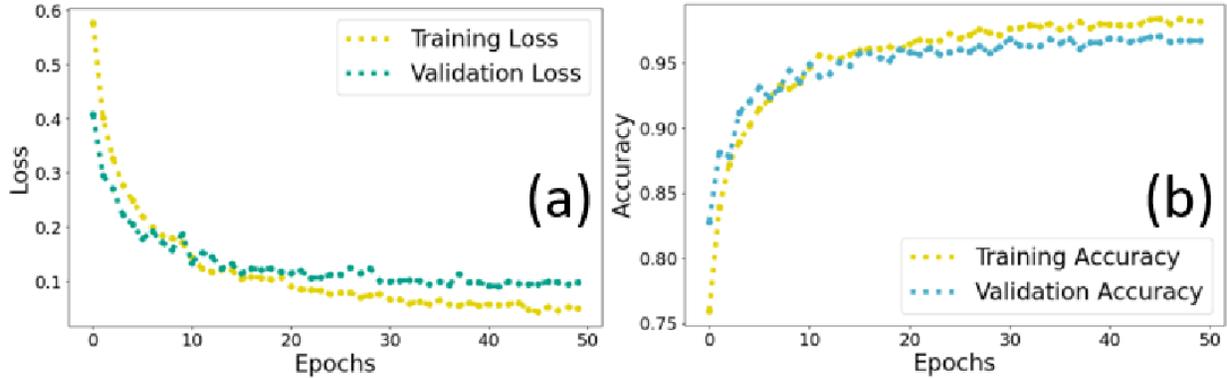

**Figure 8. Training performance of the MambaAttention model: (a) Loss curves; (b) Accuracy curves**

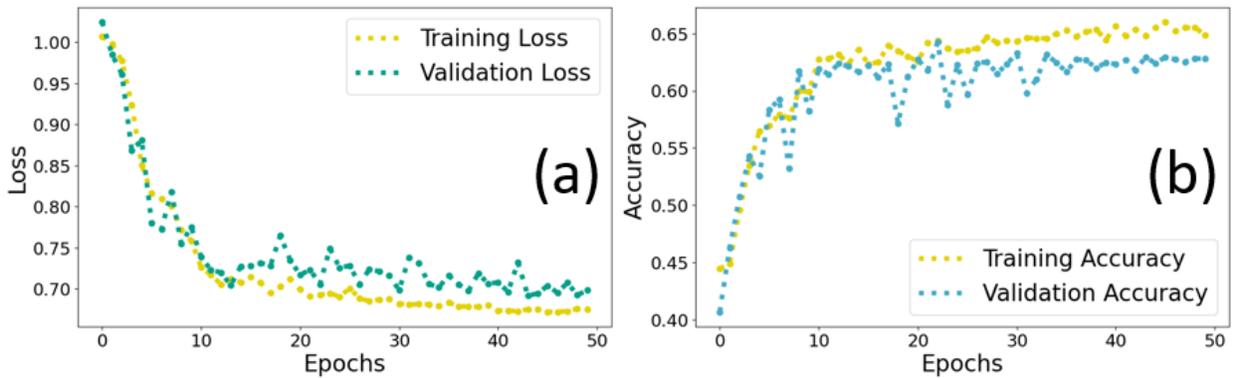

**Figure 9. Training performance of the TabTranformer model: (a) Loss curves; (b) Accuracy curves**

To further evaluate the classification performance of the models across all SAE automation levels, Receiver Operating Characteristic (ROC) curves were plotted for each model, as shown in **Figure 10**, and the corresponding Area Under the Curve (AUC) scores were computed. The MambaAttention model (**Figure 10a**) demonstrated near-perfect separation between classes, with AUC values of 0.99, 1.00, and 1.00 for Assisted Driving (Class 1), Partial Automation (Class 2), and Advanced Automation (Class 3), respectively. Its macro-AUC score of 0.9958 indicates exceptional overall discriminative capability across all categories, reinforcing the model's consistent superiority as observed in precision, recall, and F1-score metrics.

In contrast, the TabTransformer model (**Figure 10b**) showed comparatively weaker performance, particularly for Class 3 (Advanced Automation), which achieved an AUC of only 0.84. While Class 1 and Class 2 reached respectable AUC values of 0.91 and 0.74, respectively, the macro-AUC score of 0.8220 reflects the model's overall limitations in reliably distinguishing between automation levels. This aligns with earlier findings from the confusion matrix and classification report, where TabTransformer exhibited reduced performance and high misclassification rates in the mid- and high-level automation classes.

The TabPFN model (**Figure 10c**) demonstrated strong and well-balanced classification capabilities, achieving class-wise AUC scores of 0.99 (Assisted Driving), 0.99 (Partial Automation), and 1.00 (Advanced Automation). With a macro-AUC score of 0.9896, TabPFN closely followed MambaAttention in performance, confirming its robustness in detecting patterns across automation levels despite relying on a zero-shot inference paradigm without task-specific training. These ROC and AUC results validate the earlier evaluation metrics and reinforce the reliability of MambaAttention and TabPFN for practical applications in SAE-level crash classification.

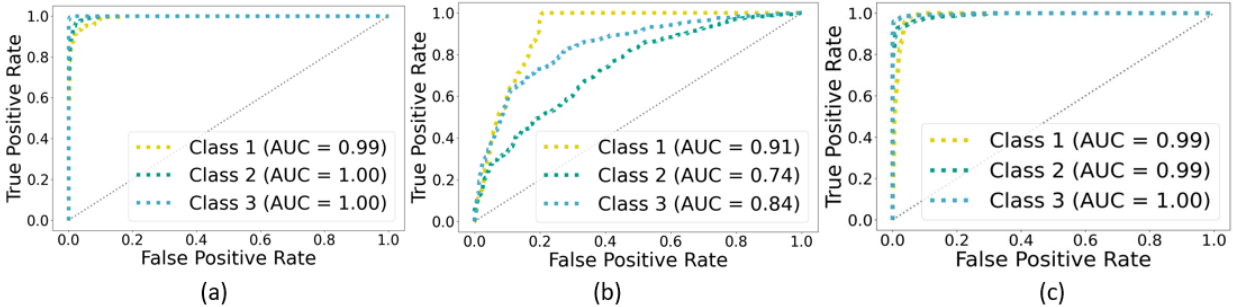

**Figure 10. ROC curve of the (a) MambaAttention, (b) TabTranformer, (c) TabPFN** (Here: Class 1 = Assisted Driving; Class 2 = Partial Automation; Class 3 = Advanced Automation)

In summary, the comparative evaluation of MambaAttention, TabTransformer, and TabPFN models for classifying crash events across SAE automation levels reveals significant insights into their respective capabilities and limitations. MambaAttention demonstrated exceptional performance, particularly in accurately identifying Advanced Automation crashes, which are often underrepresented in datasets. This aligns with studies indicating that 94% of Level 4 crashes occur in urban settings (Ding et al., 2024b), necessitating precise detection for targeted analysis of factors such as pedestrian interactions and sensor limitations. TabPFN exhibited strong zero-shot performance, enabling rapid deployment without task-specific training. Its robustness in detecting patterns across automation levels makes it a valuable tool for analyzing emerging crash patterns, including non-motorist collisions.

Conversely, TabTransformer's comparatively lower performance in classifying Partial Automation crashes highlights challenges in modeling scenarios where human drivers share control with automated systems. Given that 67% of Level 2 crashes occur on highways (Ding et al., 2024b) and often involve disengagement failures or overreliance on ADAS, misclassification risks obscuring critical human-factor variables that contribute to crash severity (Sharma et al., 2021).

These findings have significant implications for transportation safety and policy. Accurate classification of crashes by automation level is crucial for understanding the safety performance of different vehicle technologies. The superior performance of MambaAttention and TabPFN supports the NHTSA mandate for detailed Level 4 crash reporting, aiding regulators in refining safety standards for urban autonomous vehicle deployment (NHTSA, 2022). Furthermore, these results underscore the need for improved modeling techniques and data collection methods for mid-level automation scenarios. The challenges observed with TabTransformer in classifying Partial Automation crashes emphasize the importance of granular data on human-machine interactions, where factors like driver type and system sophistication significantly affect injury severity.

In conclusion, the integration of advanced deep learning models like MambaAttention and TabPFN into crash analysis framework enhances our understanding of automation-related crash dynamics. Their application can inform the development of targeted safety measures and regulatory frameworks, ultimately contributing to the safe deployment of AV technologies.

**CONCLUSIONS**

This study addresses the critical need for classifying SAE automation levels in traffic crashes, a task that is increasingly important as the deployment of AVs continues to expand. Leveraging structured crash data from the Texas CRIS database (2024), the study categorized 4,649 crashes into three automation levels: Assisted Driving, Partial Automation, and Advanced Automation. SAE Level 0, representing fully human-driven vehicles, was excluded from the analysis due to its disproportionately large volume, comprising over 600,000 crashes. Including this class would have overwhelmed the model and skewed the learning process, given the study's primary focus on understanding patterns and risks associated with emerging automation technologies represented in Levels 1 through 5. Three advanced deep learning models, MambaAttention, TabPFN, and TabTransformer, were employed to predict the SAE automation level based on crash features. After addressing class imbalances using SMOTEENN, a balanced dataset of 7,300 samples was developed for model training and evaluation. Among the models, MambaAttention demonstrated the strongest overall performance, achieving F1-scores of 88 percent, 97 percent, and 99 percent for Assisted Driving, Partial Automation, and Advanced Automation, respectively. TabPFN also showed strong zero-shot inference capabilities, particularly for Advanced Automation crashes, with an F1-score of 99 percent. TabTransformer underperformed, especially in detecting Partial Automation crashes, where the F1-score was 55 percent, likely due to challenges in modeling human-system shared control scenarios.

The novelty of this research lies in the application of specialized tabular deep learning models to classify SAE automation levels using structured crash data, a domain where such architectures have been rarely applied. The comparative evaluation of MambaAttention, TabPFN, and TabTransformer offers new insights into crash analytics for automation, particularly in identifying underrepresented Advanced Automation cases. The high classification accuracy of MambaAttention and TabPFN facilitates more reliable post-crash investigations, supports the development of automation-specific countermeasures, and informs evolving regulatory and insurance frameworks.

Despite these promising contributions, several limitations should be noted. Although the dataset was balanced using synthetic sampling techniques, it may still contain inconsistencies in reporting, particularly within the Advanced Automation category. The limited number of real-world crashes involving higher automation levels restricts the generalizability of findings. Furthermore, the study relied on data from a single state for a single year. Applying the methodology across multiple states or using national-level datasets could improve external validity. Future research should consider integrating multimodal data sources such as naturalistic driving videos, vehicle sensor outputs, and driver behavior narratives to enrich model inputs and improve classification precision. In addition, longitudinal studies that track automation performance over time could help assess the evolving safety impact of different SAE levels. These extensions would provide a more comprehensive understanding of automation-related crash dynamics and further strengthen data-driven AV policy and safety development.

**Funding**
This research received no external funding.28

# REFERENCES

Ali, Y., Haque, M.M., Mannering, F., 2023. Assessing traffic conflict/crash relationships with extreme value theory: Recent developments and future directions for connected and autonomous vehicle and highway safety research. Anal. Methods Accid. Res. 39, 100276. https://doi.org/10.1016/j.amar.2023.100276

Angarita-Zapata, J.S., Maestre-Gongora, G., Calderín, J.F., 2021. A Case Study of AutoML for Supervised Crash Severity Prediction. Presented at the 19th World Congress of the International Fuzzy Systems Association (IFSA), 12th Conference of the European Society for Fuzzy Logic and Technology (EUSFLAT), and 11th International Summer School on Aggregation Operators (AGOP), Atlantis Press, pp. 187–194. https://doi.org/10.2991/asum.k.210827.026

Antariksa, G., Tamakloe, R., Liu, J., Das, S., 2025. Automated and Explainable Artificial Intelligence to Enhance Prediction of Pedestrian Injury Severity. IEEE Trans. Intell. Transp. Syst. 26, 5568–5584. https://doi.org/10.1109/TITS.2025.3526217

Ashraf, M.T., 2021. Identification of Crash Contributing Factors in AV Involved Crashes (M.Sc.). West Virginia University, United States -- West Virginia.

Cascetta, E., Cartenì, A., Di Francesco, L., 2022. Do autonomous vehicles drive like humans? A Turing approach and an application to SAE automation Level 2 cars. Transp. Res. Part C Emerg. Technol. 134. https://doi.org/10.1016/j.trc.2021.103499

Channamallu, S.S., Almaskati, D., Kermanshachi, S., Pamidimukkala, A., American Society of Civil Engineers, 2024. Autonomous Vehicle Safety: A Comprehensive Analysis of Crash Injury Determinants. p. pp 767-779. https://doi.org/10.1061/9780784485514.067

Donà, R., Mattas, K., Vass, S., Ciuffo, B., 2024. Experimental investigation of the multianticipation mechanism in commercial SAE level 2 automated driving vehicles and associated safety impact. Accid. Anal. Prev. 208, 107784. https://doi.org/10.1016/j.aap.2024.107784

Dong, J., Chen, S., Miralinaghi, M., Chen, T., Labi, S., 2022. Development and testing of an image transformer for explainable autonomous driving systems. J. Intell. Connect. Veh. 5, 235–249. https://doi.org/10.1108/JICV-06-2022-0021

Dong, J., Chen, S., Miralinaghi, M., Chen, T., Li, P., Labi, S., 2023. Why did the AI make that decision? Towards an explainable artificial intelligence (XAI) for autonomous driving systems. Transp. Res. Part C Emerg. Technol. 156, 104358. https://doi.org/10.1016/j.trc.2023.104358

Du, R., Chen, S., Li, Y., Ha, P.Y.J., Dong, J., Anastasopoulos, P.C., Labi, S., 2021. A Cooperative Crash Avoidance Framework for Autonomous Vehicle under Collision-Imminent Situations in Mixed Traffic Stream. https://doi.org/10.1109/ITSC48978.2021.9564937

Faria, N. de O., Merenda, C., Greatbatch, R., Tanous, K., Suga, C., Akash, K., Misu, T., Gabbard, J., 2021. The Effect of Augmented Reality Cues on Glance Behavior and Driver-Initiated Takeover on SAE Level 2 Automated-Driving. Proc. Hum. Factors Ergon. Soc. Annu. Meet. 65, pp 1342-1346. https://doi.org/10.1177/1071181321651004

Favarò, F.M., Nader, N., Eurich, S.O., Tripp, M., Varadaraju, N., 2017. Examining accident reports involving autonomous vehicles in California. PLOS ONE 12, e0184952. https://doi.org/10.1371/journal.pone.0184952

Fu, H., Ye, S., Fu, X., Chen, T., Zhao, J., 2025a. New insights into factors affecting the severity of autonomous vehicle crashes from two sources of AV incident records. Travel Behav. Soc. 38, 100934. https://doi.org/10.1016/j.tbs.2024.100934

Fu, H., Ye, S., Fu, X., Chen, T., Zhao, J., 2025b. New insights into factors affecting the severity of autonomous vehicle crashes from two sources of AV incident records. Travel Behav. Soc. 38, 100934. https://doi.org/10.1016/j.tbs.2024.100934

Garbacik, N., Mastory, C., Nguyen, H., Yadav, S., Llaneras, R., McCall, R., 2021. Lateral Controllability for Automated Driving (SAE Level 2 and Level 3 Automated Driving Systems). Presented at the SAE Technical Paper, Society of Automotive Engineers (SAE). https://doi.org/10.4271/2021-01-0864

Garg, A., Das, S.S., Ramamurthi, N., 2023. AutoML in Drug Discovery: Side-Effects Prediction Using AutoGluon Framework and Its Applications in Drug Discovery, in: Proceedings of the 14th ACM International Conference on Bioinformatics, Computational Biology, and Health Informatics, BCB '23. Association for Computing Machinery, New York, NY, USA, p. 1. https://doi.org/10.1145/3584371.3613051

Gerber, M.A., Schroeter, R., Ho, B., 2023. A human factors perspective on how to keep SAE Level 3 conditional automated driving safe. Transp. Res. Interdiscip. Perspect. 22, 100959. https://doi.org/10.1016/j.trip.2023.100959

Gomes, L., 2014. Hidden Obstacles for Google's Self-Driving Cars [WWW Document]. MIT Technol. Rev. URL https://www.technologyreview.com/2014/08/28/171520/hidden-obstacles-for-googles-self-driving-cars/ (accessed 4.2.25).

Joshi, A., 2018. Hardware-in-the-Loop (HIL) Implementation and Validation of SAE Level 2 Automated Vehicle with Subsystem Fault Tolerant Fallback Performance for Takeover Scenarios. SAE Int. J. Connect. Autom. Veh. 1, pp 13-32. https://doi.org/10.4271/12-01-01-0002

Knoop, V.L., Wang, M., Wilmink, I., Hoedemaeker, D.M., Maaskant, M., Van der Meer, E.-J., 2019. Platoon of SAE Level-2 Automated Vehicles on Public Roads: Setup, Traffic Interactions, and Stability. Transp. Res. Rec. J. Transp. Res. Board 2673, pp 311-322. https://doi.org/10.1177/0361198119845885

Kuo, P.-F., Hsu, W.-T., Lord, D., Putra, I.G.B., 2024. Classification of autonomous vehicle crash severity: Solving the problems of imbalanced datasets and small sample size. Accid. Anal. Prev. 205, 107666. https://doi.org/10.1016/j.aap.2024.107666

Li, J., Liu, J., Liu, P., Qi, Y., 2020. Analysis of Factors Contributing to the Severity of Large Truck Crashes. Entropy 22, 1191. https://doi.org/10.3390/e22111191

Li, P., Chen, S., Yue, L., Xu, Y., Noyce, D.A., 2024. Analyzing relationships between latent topics in autonomous vehicle crash narratives and crash severity using natural language processing techniques and explainable XGBoost. Accid. Anal. Prev. 203, 107605. https://doi.org/10.1016/j.aap.2024.107605

Liu, P., Guo, Y., Liu, Pan, Ding, H., Cao, J., Zhou, J., Feng, Z., 2024. What can we learn from the AV crashes? – An association rule analysis for identifying the contributing risky factors. Accid. Anal. Prev. 199, 107492. https://doi.org/10.1016/j.aap.2024.107492

Liu, Q., Wang, X., Liu, S., Yu, C., Glaser, Y., 2024. Analysis of pre-crash scenarios and contributing factors for autonomous vehicle crashes at intersections. Accid. Anal. Prev. 195, 107383. https://doi.org/10.1016/j.aap.2023.107383
30

Miele, D., Ferraro, J., Mouloua, M., 2021. Driver Confidence and Level of Automation Influencing Trust in Automated Driving Features. Proc. Hum. Factors Ergon. Soc. Annu. Meet. 65, pp 1312-1316. https://doi.org/10.1177/1071181321651300

NHTSA, 2022. NHTSA Early Estimates Show Overall Increase in Roadway Deaths in First Half of 2022, Second Quarter 2022 Projects First Decline Since 2020 | NHTSA [WWW Document]. URL https://www.nhtsa.gov/press-releases/early-estimates-traffic-fatalities-first-half-2022 (accessed 3.28.25).

Nordhoff, S., Stapel, J., He, X., Gentner, A., Happee, R., 2021. Perceived safety and trust in SAE Level 2 partially automated cars: Results from an online questionnaire. PLoS One 16, e0260953. https://doi.org/10.1371/journal.pone.0260953

Olofsson, B., Nielsen, L., 2021. Using Crash Databases to Predict Effectiveness of New Autonomous Vehicle Maneuvers for Lane-Departure Injury Reduction. IEEE Trans. Intell. Transp. Syst. 22, pp 3479-3490. https://doi.org/10.1109/TITS.2020.2983553

Ostermaier, I., Gwehenberger, J., Borrack, M., Feldhütter, A., Pschenitza, M., 2019. Analyse von Unfallschäden zur Ermittlung des Unfallvermeidungspotenzials durch automatisierte Fahrfunktionen auf SAE-Level 3 und 4. Anal. Accid. Damage Determine Accid. Avoid. Potential Autom. Driv. Funct. SAE Levels 3 4 57, pp 144-50.

Penmetsa, P., Sheinidashtegol, P., Musaev, A., Adanu, E.K., Hudnall, M., 2021. Effects of the autonomous vehicle crashes on public perception of the technology. IATSS Res. 45, 485–492. https://doi.org/10.1016/j.iatssr.2021.04.003

Qi, W., Xu, C., Xu, X., 2021. AutoGluon: A revolutionary framework for landslide hazard analysis. Nat. Hazards Res. 1, 103–108. https://doi.org/10.1016/j.nhres.2021.07.002

Souders, D., Agrawal, S., Peeta, S., Center for Connected and Automated Transportation, Purdue University, Office of the Assistant Secretary for Research and Technology, 2022. Impacts of In-vehicle Alert Systems on Situational Awareness and Driving Performance in SAE Level 3 Vehicle Automation (Digital/other). https://doi.org/10.5703/1288284317573

Teoh, E.R., 2020. What's in a Name? Drivers' Perceptions of the Use of Five SAE Level 2 Driving Automation Systems. J. Safety Res. 72, pp 145-151. https://doi.org/10.1016/j.jsr.2019.11.005

Teoh, E.R., Kidd, D.G., 2017. Rage against the machine? Google's self-driving cars versus human drivers. J. Safety Res. 63, 57–60. https://doi.org/10.1016/j.jsr.2017.08.008

TESLA, 2025. Tesla Investor Relations [WWW Document]. URL https://ir.tesla.com/ (accessed 3.28.25).

Varotto, S.F., Mons, C., Hogema, J.H., Christoph, M., van Nes, N., Martens, M.H., 2022. Do adaptive cruise control and lane keeping systems make the longitudinal vehicle control safer? Insights into speeding and time gaps shorter than one second from a naturalistic driving study with SAE Level 2 automation. Transp. Res. Part C Emerg. Technol. 141, 103756. https://doi.org/10.1016/j.trc.2022.103756

Wang, J., Xue, Q., Zhang, C.W.J., Wong, K.K.L., Liu, Z., 2024a. Explainable coronary artery disease prediction model based on AutoGluon from AutoML framework. Front. Cardiovasc. Med. 11. https://doi.org/10.3389/fcvm.2024.1360548

Wang, J., Xue, Q., Zhang, C.W.J., Wong, K.K.L., Liu, Z., 2024b. Explainable coronary artery disease prediction model based on AutoGluon from AutoML framework. Front. Cardiovasc. Med. 11. https://doi.org/10.3389/fcvm.2024.1360548

Wang, S., Li, Z., 2019. Roadside Sensing Information Enabled Horizontal Curve Crash Avoidance System Based on Connected and Autonomous Vehicle Technology. Transp.

Res. Rec. J. Transp. Res. Board 2673, pp 49-60. https://doi.org/10.1177/0361198119837957

Wang, X., Peng, Y., Xu, T., Xu, Q., Wu, X., Xiang, G., Yi, S., Wang, H., 2022. Autonomous driving testing scenario generation based on in-depth vehicle-to-powered two-wheeler crash data in China. Accid. Anal. Prev. 176, 106812. https://doi.org/10.1016/j.aap.2022.106812

WHO, 2022. Road traffic injuries [WWW Document]. URL https://www.who.int/news-room/fact-sheets/detail/road-traffic-injuries (accessed 3.28.25).

Winkelman, Z., Buenaventura, M., Anderson, J., Beyene, N., Katkar, P., Baumann, G., 2019. When Autonomous Vehicles Are Hacked, Who Is Liable? RAND Corporation.

Yin, X., Liu, Q., Pan, Y., Huang, X., Wu, J., Wang, X., 2021. Strength of Stacking Technique of Ensemble Learning in Rockburst Prediction with Imbalanced Data: Comparison of Eight Single and Ensemble Models. Nat. Resour. Res. 30, 1795–1815. https://doi.org/10.1007/s11053-020-09787-0

Zhou, R., Huang, H., Lee, J., Huang, X., Chen, J., Zhou, H., 2023. Identifying typical pre-crash scenarios based on in-depth crash data with deep embedded clustering for autonomous vehicle safety testing. Accid. Anal. Prev. 191, 107218. https://doi.org/10.1016/j.aap.2023.107218

Zhou, R., Lin, Z., Zhang, G., Huang, H., Zhou, H., Chen, J., 2024. Evaluating Autonomous Vehicle Safety Performance Through Analysis of Pre-Crash Trajectories of Powered Two-Wheelers. IEEE Trans. Intell. Transp. Syst. 25, pp 13560-13572. https://doi.org/10.1109/TITS.2024.3392673

Zhu, S., Meng, Q., 2022. What can we learn from autonomous vehicle collision data on crash severity? A cost-sensitive CART approach. Accid. Anal. Prev. 174, 106769. https://doi.org/10.1016/j.aap.2022.106769